\documentclass[conference]{IEEEtran}
\IEEEoverridecommandlockouts
\usepackage{cite}
\usepackage{amsmath,amssymb,amsfonts}
\usepackage{graphicx}
\usepackage{textcomp}
\usepackage{xcolor}
\usepackage{hyperref}
\usepackage{cleveref}
\usepackage{algorithm}
\usepackage{algpseudocode}
\usepackage{booktabs}
\usepackage{multirow}
\usepackage[most]{tcolorbox}
\usepackage{xspace}
\usepackage{makecell}
\def\BibTeX{{\rm B\kern-.05em{\sc i\kern-.025em b}\kern-.08em
    T\kern-.1667em\lower.7ex\hbox{E}\kern-.125emX}}

\def \alg{\texttt{SpecMind}\xspace}
\def \algg{\texttt{SpecBench}\xspace}

\ifodd 1
\newcommand{\congr}[1]{{\color{red}#1}}
\else
\newcommand{\congr}[1]{#}
\fi

\ifodd 0
\newcommand{\congc}[1]{{\color{red}(Cong: #1)}}
\else
\newcommand{\congc}[1]{}
\fi

\begin{document}

\title{\alg: Enabling Spectrum Intelligence via Multi-Agent Hybrid Retrieval-Augmented Generation\\
\thanks{This work was supported in part by SpectrumX, the National Science Foundation (NSF) Spectrum Innovation Center, through grant AST 2132700 operated under Cooperative Agreement by the University of Notre Dame.}
\thanks{\(^*\)The first two authors contributed equally to this work.}
}

\author{\IEEEauthorblockN{Songwei Dong\(^*\)\(^\ddagger\), Bingyan Lu\(^*\)\(^\dagger\), Makayla Kienlen\(^\S\), J. Nicholas Laneman\(^\dagger\) and Cong Shen\(^\ddagger\)}
\IEEEauthorblockA{
\textit{ \(^\ddagger\) Department of
Electrical and Computer Engineering, University of Virginia} \\
\textit{\(^\dagger\)Department of Electrical Engineering, University of Notre Dame}\\ 
\textit{\(^\S\)Baskin School of Engineering, University of California, Santa Cruz}\\
\(^\ddagger\) \{hxt5ap, cong\}@virginia.edu, \(^\dagger\)\{blu, jnl\}@nd.edu, \(^\S\) mkienlen@ucsc.edu
}
}

\maketitle

\begin{abstract}
The exponential growth of wireless devices is driving unprecedented spectrum demand, pushing spectrum management toward more fine-grained decisions across space, time, and device constraints. As a result, spectrum policymakers and engineers must process large volumes of data that come from diverse sources and take many different forms, such as text and tables. These data sources are often disaggregated and require significant time and effort to integrate, search, and interpret. Furthermore, most of this information is formatted for human understanding and is not readily accessible to automated systems. To address this challenge, we propose \alg, a novel Multi-Agent Retrieval-Augmented Generation (RAG) system for spectrum intelligence that performs reasoning over heterogeneous data sources. This system enables autonomous agents to coordinate specialized sub-agents that retrieve and synthesize knowledge across policy proceedings, legal regulations, and license databases. We develop \algg, a question and answer (Q\&A) dataset based on real-world license records and policy proceedings, addressing the lack of evaluation resources for RAG systems in the spectrum domain. Experimental results demonstrate that \alg outperforms traditional, general-purposed RAG systems across spectrum-related tasks, achieving over 80\% win rate against strong baselines. The agent-based design enables more accurate retrieval, better contextual reasoning, and improved task completion across diverse query types.

\end{abstract}


\section{Introduction}
\nocite{RAG_in_Spectrum_Policy}

The wireless spectrum faces increasing demand as modern services and devices compete for limited bandwidth. Spectrum management has consequently become more complex, involving constraints across frequency, time, location, users, and regulatory requirements~\cite{marshall2017three}. This complexity drives rapid growth in spectrum data in both scale and heterogeneity. Routine tasks such as transmission compliance verification and spectrum policy development require integrating fragmented, heterogeneous information. These challenges necessitate scalable and context-aware methods for spectrum data interpretation and decision support~\cite{rutagemwa2024accelerating}.

Large language models (LLMs) have demonstrated strong capabilities in domain-specific understanding and reasoning~\cite{bariah2023understanding,mann2020language}, while retrieval-augmented generation (RAG) improves grounding by incorporating external knowledge sources~\cite{singh2025agentic, gao2023retrieval, guu2020realmretrievalaugmentedlanguagemodel}. Recent work has begun to extend RAG to spectrum and telecommunications domains. SpectrumRAG~\cite{lu2025initial} introduces an iterative framework that leverages query rewriting to refine retrieval. TelcoRAG~\cite{bornea2024telco} enhances retrieval over telecommunications standards through query augmentation and specialized pipelines. Radio Regulations GPT~\cite{kassimi2025retrieval} proposes a domain-specific RAG pipeline for regulatory understanding. However, these approaches remain largely centered on text-based retrieval and lack adaptive mechanisms for integrating heterogeneous, multi-source data, limiting their effectiveness on increasingly complex tasks.

\congc{cite the AI4NextG workshop paper.}

To address these limitations, we propose \alg, an Agentic RAG framework for spectrum intelligence over heterogeneous data. Our key contributions include the construction of a spectrum-specific corpus, the design of specialized tools and prompts, and the development of an agentic framework that orchestrates these components for complex reasoning and task execution that supports tasks such as incumbent investigation, stakeholder analysis, and regulatory compliance verification. We further introduce a benchmark for joint license and proceeding tasks and conduct a systematic evaluation of \alg. Our contributions are summarized as follows:
\begin{itemize}
\item \textbf{Reusable, extensible spectrum knowledge databases:}
We construct a unified corpus of FCC licensing records, proceeding documents, and regulatory texts as complementary machine-readable resources. Licensing data are organized in a relational SQL database, proceedings are modeled as graphs capturing entities and cross-document relations, and regulatory texts are indexed via embeddings for retrieval. The resulting corpus is reusable and extensible, enabling efficient integration of new records.

\item \textbf{An expert-designed benchmark for heterogeneous spectrum-focused RAG:} We introduce \algg, an expert-designed benchmark dataset for evaluating RAG systems over heterogeneous spectrum data. It comprises 450 curated question–answer pairs reflecting realistic spectrum analysis tasks, including fact verification, cross-source synthesis, and multi-hop reasoning. Built upon the proposed databases, \algg enables standardized and reproducible evaluation and serves as a valuable resource for spectrum-oriented RAG.

\item \textbf{A multi-agent hybrid RAG system for heterogeneous sources:} To the best of our knowledge, \alg is the first multi-agent hybrid RAG framework for spectrum intelligence that enables coordinated retrieval and reasoning over heterogeneous spectrum data, including tabular, graph-structured, and textual sources. By supporting modality-aware retrieval and cross-source reasoning, \alg improves response accuracy and completeness over existing spectrum-oriented RAG pipelines, achieving over 80\% win rate against strong baselines.
\end{itemize}

We have publicly released the \algg dataset, the reusable spectrum databases, and the \alg framework implementation to support reproducibility and community research~\cite{RAG_in_Spectrum_Policy}.

\section{Methodology}
\subsection{Database Construction}

\begin{figure}[!t]
    \centering
    \includegraphics[width=0.9\columnwidth]{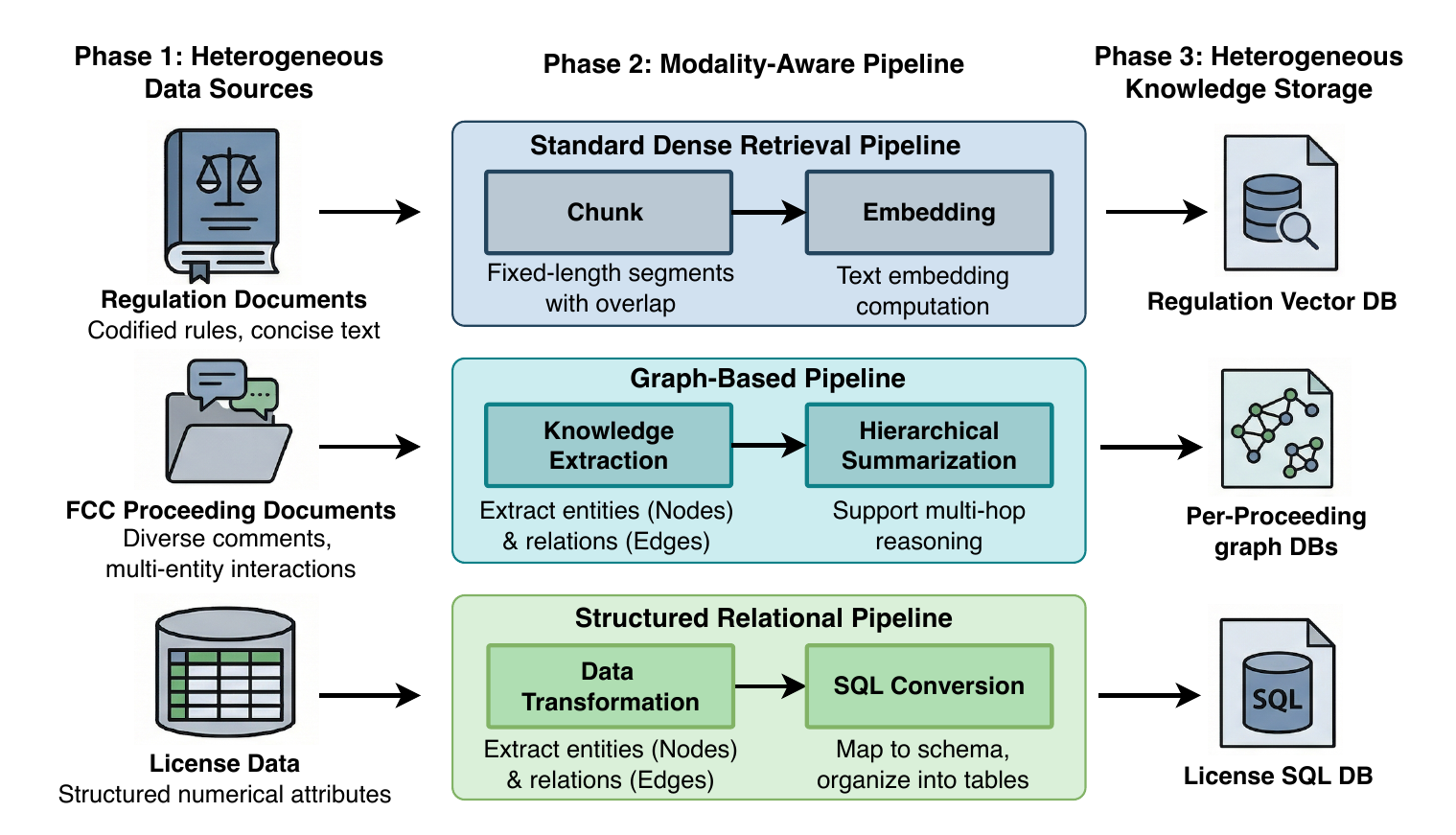}
    \caption{Database construction pipeline for heterogeneous spectrum data.}
    \label{fig:database}
\end{figure}

The construction process is illustrated in \Cref{fig:database}. We consider three primary data sources in the spectrum domain: (1) \textit{license data}, which appear as structured tables with numerical attributes; (2) \textit{proceeding documents}, released by the FCC, each centered on a specific topic and containing diverse comments from individuals and organizations during regulatory deliberation; and (3) \textit{regulatory documents}, which codify existing rules and policies. Applying RAG to such heterogeneous sources requires constructing modality-aware databases tailored to their distinct characteristics, rather than relying on a general-purposed semantic retrieval pipeline.

For \textit{regulatory texts}, we adopt a standard dense retrieval paradigm. Documents are segmented into fixed-length chunks (1000 tokens with 200-token overlap) to preserve semantic continuity. Given a query, we retrieve top-$k$ candidates via embedding similarity and further apply a reranking stage to refine relevance. Specifically, we select the top 20 candidates from initial retrieval and rerank them to obtain the top 5 final contexts. This design aligns with the nature of regulatory texts, which are concise and legally precise, making queries rely heavily on accurate semantic matching.

In contrast, \textit{FCC proceeding} comments and reply comments exhibit high volume, topic diversity, and complex multi-entity interactions. A single entity may express inconsistent positions across proceedings, and many queries require cross-document synthesis (e.g., identifying agency stances or comparing stakeholder views). Such requirements are poorly served by chunk-level semantic retrieval. We therefore enhance GraphRAG~\cite{edge2025localglobalgraphrag}, which models documents as graphs with entities as nodes and relations as edges, and construct hierarchical summaries to support multi-hop reasoning. We build a separate graph database for each proceeding to ensure structural consistency (e.g., avoiding cross-document conflicts introduced by merging multiple proceedings into a single graph) and enable modular extensibility without rebuilding the entire corpus when new proceedings are introduced. Detailed statistics of the graph corpora, including node distributions across hierarchy levels, are summarized in \Cref{tab:graph_corpora}.

\textit{License data} present a fundamentally different challenge, as they are dominated by structured numerical fields with weak semantic signals~\cite{wallace-etal-2019-nlp}. Traditional RAG methods perform poorly in this setting due to their reliance on semantic similarity. We instead model license data using relational databases and perform retrieval via SQL queries to enable exact matching over structured attributes. The problem is thus reformulated as generating executable SQL queries from natural language questions, which can be effectively handled by LLMs. Concretely, we extract license records into intermediate JSON representations for preprocessing, and then convert them into a relational SQL database. We further organize the database into service-specific subtables to accommodate heterogeneous schemas across different license types.

\begin{table}[ht]
\caption{Graph corpus construction results for FCC proceedings. NTIA NSS denotes the NTIA National Spectrum Strategy.}
\label{tab:graph_corpora}
\begin{center}
\begin{tabular}{lccccc}
\toprule
\multirow{2}{*}{\bf Proceeding} & \multicolumn{5}{c}{\bf Number of Nodes} \\ 
\cmidrule(lr){2-6}
& \textbf{Level 0} & \textbf{Level 1} & \textbf{Level 2} & \textbf{Level 3} & \textbf{Level 4} \\ 
\midrule
FCC 19 - 38   & 477 & 445 & 133 & 57 & - \\
FCC 24 - 72   & 375  & 329  & 163 & 23 & - \\
FCC 25 - 59   & 1045  & 1008  & 419 & 40 & - \\
FCC 22 - 352  & 2094 & 2051  & 1599  & 571 & - \\
FCC 23 - 158  & 902  & 868  & 405 & 53 & - \\
FCC 23 - 232  & 871  & 794  & 338 & 82 & 58 \\
NTIA NSS & 4079 & 3960 & 3074 & 679 & 345 \\
\bottomrule
\end{tabular}
\end{center}
\end{table}

\subsection{Framework Design}

\begin{figure*}[t]
\centering
\includegraphics[width=0.95\textwidth]{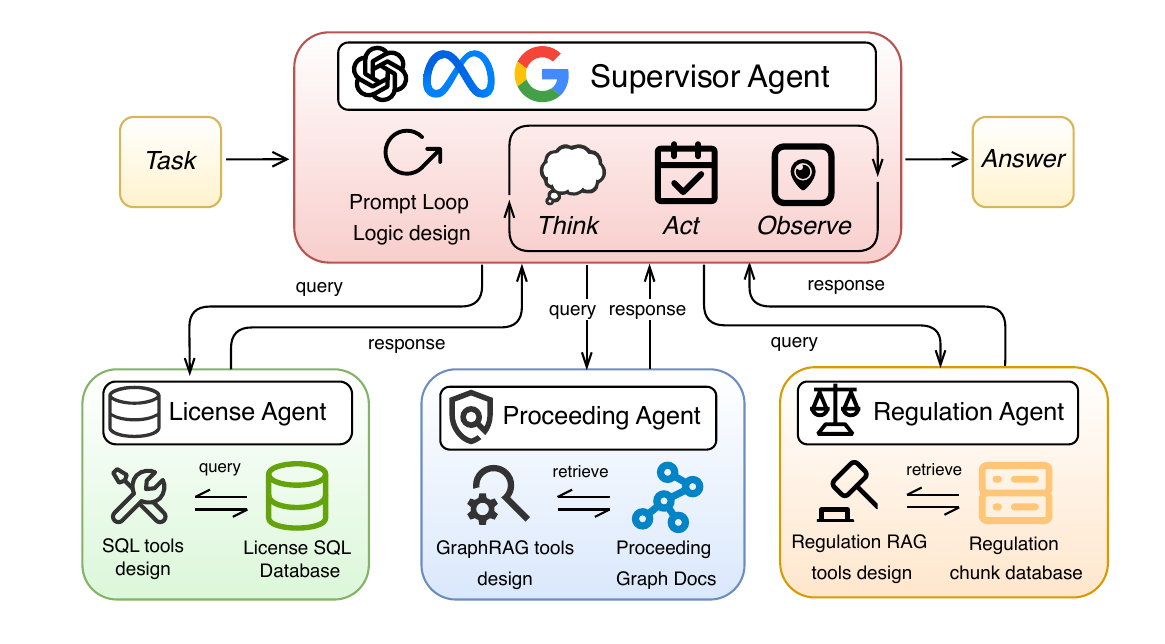}
\caption{Overview of the proposed {\alg} framework.}
\label{fig:overview}
\end{figure*}

We propose \alg, a multi-agent hybrid RAG system
for reasoning over heterogeneous spectrum data. The overall system design is illustrated in \Cref{fig:overview}. 
Each task instance is represented as
\[
(Q, \mathcal{K}, \mathcal{A}, \Pi, \mathcal{T}, A),
\]
where $Q$ denotes a user query, $A$ is the ground-truth answer, and $\mathcal{K}$ comprises heterogeneous knowledge sources, including structured license databases, graph-structured proceeding documents, and unstructured regulatory texts. The agent set is defined as
\[
\mathcal{A} = \{a_{\text{sup}}\} \cup \mathcal{A}_{\text{sub}},
\quad
\mathcal{A}_{\text{sub}} = \{a_{\text{lic}}, a_{\text{proc}}, a_{\text{reg}}\}.
\]
Each agent $a \in \mathcal{A}$ is associated with an action policy $\pi_a \in \Pi$. The Supervisor Agent $a_{\text{sup}}$ coordinates task decomposition, agent invocation, and result integration, while specialized agents perform domain-specific actions.

Given a query $Q$, the system maintains a global task state $\mathcal{T}$
that accumulates intermediate evidence and partial results across multiple
agent interactions. The Supervisor Agent coordinates the execution of
specialized agents and iteratively refines the task state until a final answer
$\hat{A}$ is produced.

The objective of the system is to generate an answer $\hat{A}$ that is
factually grounded in $\mathcal{K}$ and consistent with the ground-truth
answer $A$, while enabling accurate and flexible reasoning across
heterogeneous spectrum data sources.

\begin{algorithm}[t]
\caption{Supervisor Agent Control Loop}
\label{alg:supervisor}
\begin{algorithmic}[1]
\Require User query $Q$, knowledge sources $K$, agent set $\mathcal{A} = \{a_{\text{sup}}\} \cup \mathcal{A}_{\text{sub}}$, agent policies $\Pi$
\Ensure Final answer $\hat{A}$
\State Initialize global task state $\mathcal{T}_0 \gets \emptyset$
\State Set termination flag $\texttt{done} \gets \textbf{false}$
\State Set iteration index $t \gets 1$
\While{\textbf{not} $\texttt{done}$}
    \State \textbf{Think:}
    \State \quad Determine next action $u_t \sim \pi_{\text{sup}}(Q, \mathcal{T}_{t-1})$
    \State \textbf{Act:}
    \If{$u_t$ invokes a specialized agent $a \in \mathcal{A}_{\text{sub}}$}
        \State \quad Execute $a$ using its policy $\pi_a$ and tools, obtain result $r_t$
    \Else
        \State \quad Execute internal operation (e.g., subtask rewriting or evidence extraction), obtain result $r_t$
    \EndIf
    \State \textbf{Observe:}
    \State \quad Update task state $\mathcal{T}_t \gets \mathcal{T}_{t-1} \cup \{r_t\}$
    \State \quad Update $\texttt{done}$ based on $\mathcal{T}_t$
    \State \quad $t \gets t + 1$
\EndWhile
\State Aggregate evidence in $\mathcal{T}_{t-1}$ to generate final answer $\hat{A}$
\State \Return $\hat{A}$
\end{algorithmic}
\end{algorithm}

\subsection{Prompt Engineering}

\textbf{Design principle.} We formulate prompt engineering as the explicit specification of agent-level action policies $\Pi = \{\pi_{\text{sup}}, \pi_{\text{lic}}, \pi_{\text{proc}}, \pi_{\text{reg}}\}$, which govern task decomposition, tool usage, and cross-agent coordination. Inspired by the ReAct paradigm~\cite{yao2023reactsynergizingreasoningacting}, which interleaves reasoning and action, our design extends this principle to a multi-agent setting tailored to spectrum-specific tasks. Rather than relying on implicit reasoning in LLMs, we enforce structured decision-making through prompt-level control, enabling reliable interaction with heterogeneous data sources.

\textbf{Supervisor prompt.} The Supervisor Agent is prompted to follow a structured \textit{Think--Act--Observe} loop for multi-step reasoning. The prompt explicitly instructs the agent to decompose the query into subtasks, decide whether to invoke a specialized agent or perform an internal operation, and iteratively update the global task state. This design enables adaptive planning and dynamic routing across heterogeneous tools, while ensuring that intermediate results are incorporated into subsequent reasoning steps.

\textbf{License agent prompt.} For structured license data, we design a tool-aware prompt that explicitly specifies the usage of four SQL tools, including schema inspection, query execution, and query validation. The action policy $\pi_{\text{lic}}$ enforces a stepwise interaction pattern, where the agent first inspects database structure, then generates executable SQL queries, and finally refines results based on returned outputs. This design enables precise handling of numerical and table-centric queries beyond the capability of semantic retrieval.

\textbf{Proceeding agent prompt.} For graph-structured proceeding data, the prompt encodes a hierarchical retrieval strategy. The action policy $\pi_{\text{proc}}$ first performs proceeding selection via topic matching, and then dynamically chooses among three GraphRAG tools (basic, local, and global) according to query granularity. Specifically, local retrieval is preferred for entity-centric queries, whereas global retrieval is used for higher-level summarization. To improve robustness under incomplete graph construction, the prompt further incorporates a fallback rule that invokes basic retrieval when structured retrieval fails to return sufficient evidence.

\textbf{Regulation agent prompt.} For regulatory documents, we design a minimal high-precision prompt that focuses on retrieval refinement rather than tool selection. The action policy $\pi_{\text{reg}}$ adopts a two-stage retrieve-and-rerank pipeline, where embedding-based retrieval is followed by lightweight reranking with \texttt{Qwen3-Reranker-0.6B}. This design improves evidence precision for regulation-oriented queries, where correctness is critical.

\section{\algg Dataset}

We introduce \algg, a real-world benchmark dataset for evaluating RAG systems in the spectrum domain. \algg addresses the lack of benchmarks for heterogeneous spectrum data sources, including proceedings, license databases, and regulations. \algg adopts a question and answer (Q\&A) format, a standard paradigm for evaluating both retrieval accuracy and generation faithfulness~\cite{bansal2025can,kassimi2025retrieval}.

Following established RAG evaluation dimensions~\cite{chen2024benchmarking}, we consider three core capabilities: (i) \textit{noise robustness}, i.e., the ability to identify correct evidence under noisy retrieval results; (ii) \textit{information integration}, i.e., the ability to synthesize evidence across multiple sources; and (iii) \textit{negative rejection}, i.e., the ability to detect insufficient supporting evidence and abstain from answering, thereby avoiding hallucinated responses.

To unify evaluation across modalities, we define an \textit{evidence unit} as the minimal retrievable element, corresponding to a document for unstructured data and a table cell for structured data. A question is classified as \textit{single-source} if it can be answered using one evidence unit, and \textit{multi-source} otherwise. This distinction enables controlled evaluation of noise robustness and information integration.

\begin{table}[ht]
\centering
\caption{Overview of \algg question categories and evaluated capabilities.}
\label{tab:specbench_overview}
\begin{tabular}{l l l l}
\toprule
\textbf{Category} & \textbf{Subtype} & \textbf{Data Source} & \textbf{Evaluated Capability} \\
\midrule
Proceeding & Single-cell & \multirow{2}{*}{Proceedings} & Noise Robustness \\
& Multi-cells  &                                        & Information Integration \\
\midrule
License    & Single-cell & \multirow{2}{*}{Licenses} & Noise Robustness \\
& Multi-cells  &                                 & Information Integration \\
\midrule
Regulation & WiLL~\cite{bansal2025can} & FCC Title 47 & Information Integration \\
\midrule
Compound   & Parallel   & \multirow{2}{*}{Multi-source} & \multirow{2}{*}{Information Integration} \\
& Sequential &                              & \\

\midrule
\makecell{Unanswerable} & - & - & Negative Rejection \\ 

\bottomrule
\end{tabular}
\end{table}

As summarized in Table~\ref{tab:specbench_overview}, \algg organizes questions by data source and task structure. Single-source questions primarily evaluate noise robustness, while multi-source and compound questions assess information integration. Compound questions further test cross-agent coordination and are categorized into \textit{parallel} and \textit{sequential} types. In \textit{parallel} questions, required evidence can be retrieved independently from different sources and combined at the final stage. By contrast, \textit{sequential} questions involve inter-step dependencies, where intermediate results from one retrieval step are required to formulate subsequent queries, making them more susceptible to error propagation. Unanswerable questions contain no valid supporting evidence and evaluate negative rejection.

\begin{figure}[t]
    \centering
    \includegraphics[width=0.95\columnwidth]{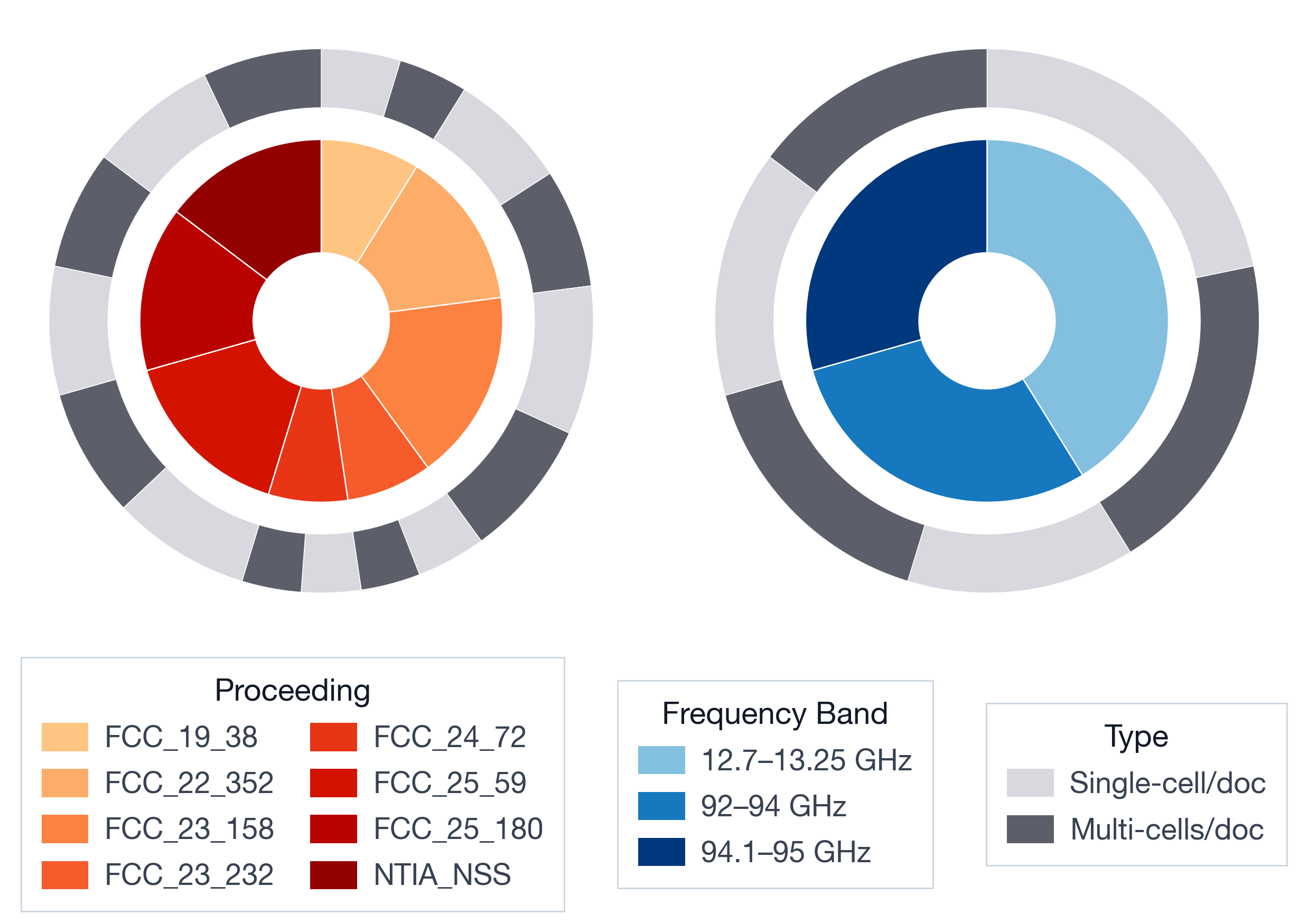}
    \caption{Distribution of questions in \algg. The inner ring shows the proportion of questions across different sources, while the outer ring distinguishes between single-source and multi-source questions.}
    \label{fig:benchmark}
\end{figure}

We construct \algg based on question types identified through real-world domain experts interviews, ensuring coverage of practical tasks. For each question, we manually retrieve supporting documents and provide a evidence-grounded reference answer. For regulation questions, we incorporate adapted Q\&A samples from the \textit{WiLL} benchmark~\cite{bansal2025can} to improve coverage of regulation-focused queries. The dataset contains 450 Q\&A pairs spanning proceeding (31.1\%), license (31.1\%), regulation (13.3\%), compound (14.4\%), and unanswerable (10.0\%), balancing realism and annotation quality. The distribution of questions across data sources and task types is illustrated in Fig.~\ref{fig:benchmark}.


\section{Experiments}

\subsection{Baselines}

\textbf{Web-search RAG.} This baseline uses Google Search as the retriever, collecting the top-20 results per query as generation context. It serves as a practical baseline for spectrum question answering using publicly accessible resources.

\textbf{SpectrumRAG.} We include SpectrumRAG~\cite{lu2025initial} as a competitive baseline, which is an iterative RAG framework that leverages LLM-based query rewriting to refine retrieval for spectrum policy question answering. This design improves retrieval quality and downstream generation, making it a strong representative of advanced RAG systems in this domain.

\begin{table*}[t]
\centering
\caption{Performance across question types under different backbones. Win: win rate; Success: success rate.}
\label{tab:main_results}
\scriptsize
\setlength{\tabcolsep}{8pt}
\renewcommand{\arraystretch}{0.9}

\begin{tabular}{ll cc cc cc cc cc cc}
\toprule
\multirow{3}{*}{Model}
& \multirow{3}{*}{Method}
& \multicolumn{12}{c}{Question Type} \\
\cmidrule(lr){3-14}
& & \multicolumn{2}{c}{Proceeding}
& \multicolumn{2}{c}{License}
& \multicolumn{2}{c}{Regulation}
& \multicolumn{2}{c}{Compound}
& \multicolumn{2}{c}{Unanswerable}
& \multicolumn{2}{c}{Overall} \\
\cmidrule(lr){3-4}
\cmidrule(lr){5-6}
\cmidrule(lr){7-8}
\cmidrule(lr){9-10}
\cmidrule(lr){11-12}
\cmidrule(lr){13-14}
& & Win & Success
& Win & Success
& Win & Success
& Win & Success
& Win & Success
& Win & Success \\
\midrule

\multirow{3}{*}{Qwen3-8B}
& Web-search RAG & 2.9 & 70.2 & 0.0 & 15.5 & 29.7 & 88.3 & 4.8 & 9.2 & - & \textbf{100} & 5.4 & 37.8 \\
& SpectrumRAG & 12.5 & 71.5 & 0.8 & 6.7 & 6.7 & 52.4 & 0.0 & 0.0 & - & \textbf{100} & 6.1 & 42.6 \\

& \textbf{\alg} 
& \textbf{60.4} & \textbf{91.8} 
& \textbf{79.3} & \textbf{87.9} 
& \textbf{42.0} & \textbf{96.7} 
& \textbf{70.5} & \textbf{81.7} 
& - & \textbf{100} 
& \textbf{66.2} & \textbf{89.8} \\

\midrule

\multirow{3}{*}{GPT-5.2}
& Web-search RAG & 4.7 & 78.3 & 0.0 & 25.3 & 38.3 & 95.8 & 8.3 & 13.3 & - & \textbf{100} & 7.9 & 50.5 \\
& SpectrumRAG & 18.8 & 78.8 & 1.2 & 8.2 & 8.3 & 60.6 & 0.0 & 0.0 & - & \textbf{100} & 8.5 & 56.5 \\
& \textbf{\alg} 
& \textbf{72.9} & \textbf{100} 
& \textbf{97.6} & \textbf{100} 
& \textbf{51.7} & \textbf{100} 
& \textbf{86.8} & \textbf{96.7} 
& - & \textbf{100} 
& \textbf{81.1} & \textbf{99.6} \\

\bottomrule
\end{tabular}
\end{table*}

\subsection{Experimental Setup}
\definecolor{lightgraybox}{RGB}{240,240,240}

\newtcolorbox{rubricbox}{
  breakable,
  colback=lightgraybox,
  colframe=black!40,
  title=\textbf{Evaluation Rubric Prompt},
  fonttitle=\bfseries,
  fontupper=\ttfamily\footnotesize,
  left=1mm,right=1mm,top=1mm,bottom=1mm
}

\textbf{Models.} To ensure a fair comparison, we evaluate all methods, including the baselines and \alg, using two backbone LLMs of different scales, Qwen3-8B and GPT-5.2. Each method uses the same backbone for both RAG and reasoning, and \texttt{text-embedding-3-small} is used for retrieval across all methods.

\textbf{Metrics.} We report \textit{success rate} and \textit{win rate} for each query. The success rate evaluates whether a method produces a valid response under one of the two settings: (i) for standard question types, including license, proceeding, regulation, and compound queries, the method must return a factually grounded answer when sufficient supporting evidence exists; and (ii) for negative-rejection questions, where no valid evidence is available, the method must correctly abstain and avoid hallucination by issuing a rejection. The win rate measures comparative answer quality. We use a strong proprietary LLM, Gemini 3.1-Pro-Preview, as a unified evaluator, which is distinct from the backbone models and fixed across all experiments to ensure consistent comparisons. For each query, the evaluator selects the best answer among all candidates; if all answers are incorrect, no method is assigned a win. 

\begin{rubricbox}
Given a question and its ground-truth answer, evaluate the responses generated by different methods.

For each response, assess:

1. Correctness: whether the answer is factually consistent with the ground truth.
2. Completeness: whether all key aspects of the question are addressed.
3. Faithfulness: whether the answer avoids hallucination and unsupported claims.

Select the response that is both correct and of the highest overall quality. If all responses are incorrect, do not select any answer.
\end{rubricbox}

\subsection{Main Results}

As shown in \Cref{tab:main_results}, \alg consistently outperforms the baselines, including Web-search RAG and SpectrumRAG, across question types under both backbone LLMs, achieving substantial win-rate gains, especially on license and compound queries, demonstrating effective structured retrieval and cross-source reasoning.

In terms of success rate, \alg maintains strong performance across all categories, achieving near-perfect results with GPT-5.2 and consistently high accuracy with Qwen3-8B. In contrast, the baselines achieve significantly lower success rates, particularly on license and compound tasks, highlighting their limitations in precise retrieval and coordinated reasoning. While Web-search RAG performs relatively well on regulation queries due to strong semantic retrieval, it fails to generalize to structured and multi-source settings.

For unanswerable questions, all methods achieve perfect success rates. This is largely because the task primarily requires abstention rather than evidence synthesis, and modern backbone LLMs already exhibit strong intrinsic capabilities in recognizing insufficient evidence and avoiding unsupported generation. As a result, performance on this category is less sensitive to the retrieval framework and more dependent on the underlying model’s calibration.

\subsection{Ablation Studies}

\begin{table*}[t]
\centering
\caption{Ablation study: replacing each sub-agent with Naive RAG (without reranking).}
\label{tab:ablation}
\footnotesize
\setlength{\tabcolsep}{4pt}
\renewcommand{\arraystretch}{1.1}

\resizebox{\textwidth}{!}{
\begin{tabular}{l cc cc cc cc cc cc}
\toprule
\multirow{3}{*}{Method}
& \multicolumn{12}{c}{Question Type} \\
\cmidrule(lr){2-13}
& \multicolumn{2}{c}{Proc.}
& \multicolumn{2}{c}{Lic.}
& \multicolumn{2}{c}{Reg.}
& \multicolumn{2}{c}{Comp.}
& \multicolumn{2}{c}{Neg.}
& \multicolumn{2}{c}{All} \\
\cmidrule(lr){2-3}
\cmidrule(lr){4-5}
\cmidrule(lr){6-7}
\cmidrule(lr){8-9}
\cmidrule(lr){10-11}
\cmidrule(lr){12-13}
& W & S
& W & S
& W & S
& W & S
& W & S
& W & S \\
\midrule

\alg
& 72.9 & 100
& 97.6 & 100
& 51.7 & 100
& 86.8 & 96.7
& - & 100
& 81.1 & 99.6 \\

\textit{w/o License Agent (RAG)}
& 70.2 {\color{red}(-2.7)} & 98.5 {\color{red}(-1.5)}
& 9.8 {\color{red}(-87.8)} & 17.2 {\color{red}(-82.8)}
& 48.9 {\color{red}(-2.8)} & 97.6 {\color{red}(-2.4)}
& 66.5 {\color{red}(-20.3)} & 80.1 {\color{red}(-16.6)}
& - & 100{\color{red}(-0.0)}
& 66.8 {\color{red}(-14.3)} & 84.4 {\color{red}(-15.2)} \\

\textit{w/o Proceeding Agent (RAG)}
& 45.6 {\color{red}(-27.3)} & 89.7 {\color{red}(-10.3)}
& 94.8 {\color{red}(-2.8)} & 98.7 {\color{red}(-1.3)}
& 49.5 {\color{red}(-2.2)} & 97.9 {\color{red}(-2.1)}
& 63.2 {\color{red}(-23.6)} & 79.4 {\color{red}(-17.3)}
& - & 100{\color{red}(-0.0)}
& 63.1 {\color{red}(-18.0)} & 87.2 {\color{red}(-12.4)} \\

\textit{w/o Regulation Agent (RAG)}
& 71.8 {\color{red}(-1.1)} & 99.3 {\color{red}(-0.7)}
& 96.2 {\color{red}(-1.4)} & 99.1 {\color{red}(-0.9)}
& 46.7 {\color{red}(-5.0)} & 93.6 {\color{red}(-6.4)}
& 84.7 {\color{red}(-2.1)} & 95.1 {\color{red}(-1.6)}
& - & 100{\color{red}(-0.0)}
& 80.3 {\color{red}(-0.8)} & 99.0 {\color{red}(-0.6)} \\

\bottomrule
\end{tabular}
}
\end{table*}

To analyze the source of performance gains, we conduct ablation studies on the three specialized agents under the GPT-5.2 backbone by individually replacing each sub-agent with a naive RAG module (without re-ranking) while preserving the overall framework. Results are reported in Table~\ref{tab:ablation}.

We observe consistent performance degradation across all ablations, confirming that each agent contributes to the overall effectiveness. The impact is most pronounced on the corresponding question types, followed by compound queries, reflecting error propagation in multi-step reasoning.

The degradation is particularly severe for the license agent, with substantial drops in both win rate and success rate (up to $-87.8$ and $-82.8$), highlighting the importance of SQL-based retrieval for structured data. The proceeding agent also plays a critical role, with notable declines on proceeding questions ($-27.3$ win, $-10.3$ success), primarily due to failures in multi-document reasoning, underscoring the benefit of graph-based retrieval. In contrast, the regulation agent shows relatively smaller impact, as its improvements mainly arise from prompt design and reranking over text-based retrieval, which offer more limited gains than structured and graph-based mechanisms.

\section{Qualitative Analysis}
\subsection{Illustrative Example}

\begin{figure}[ht]
    \centering
    \includegraphics[width=0.95\columnwidth]{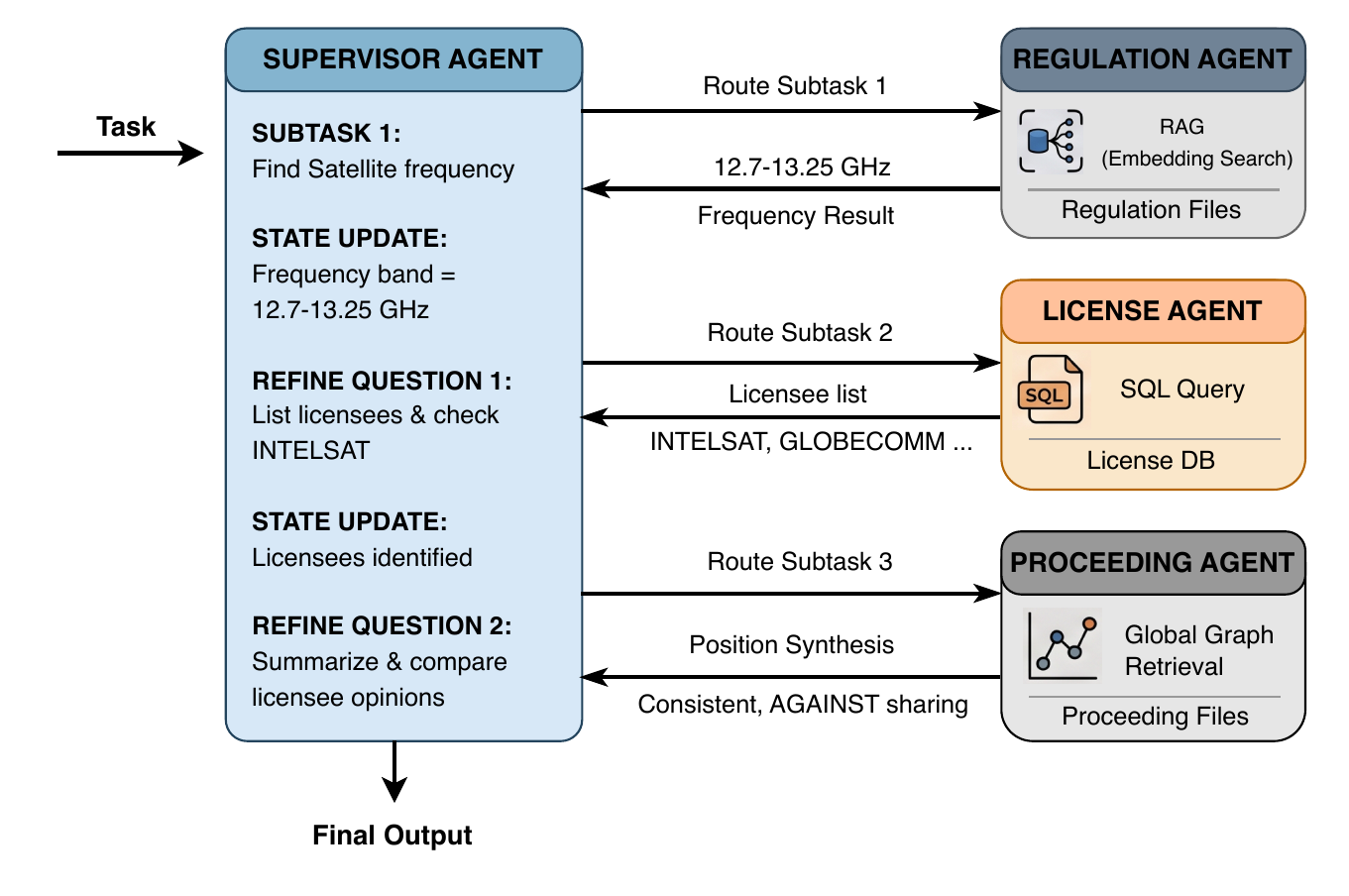}
    \caption{Case study of \alg on a representative compound query.}
    \label{fig:case_study}
\end{figure}

To illustrate how \alg operates, we present a compound Q\&A example jointly supported by all three specialized agents: \textit{Within the 12.7--13.25 GHz band, which frequencies are allocated to satellite services? Is Intelsat an incumbent in this band, and if so, do Intelsat and other incumbents share the same position on spectrum sharing with terrestrial services?} 

The reasoning process is illustrated in Fig.~\ref{fig:case_study}. The supervisor agent operates in a sequential, state-driven manner, iteratively selecting which agent to invoke and what subtask to issue. It first queries the regulation agent to identify the relevant frequency band (e.g., 12.7--13.25 GHz), establishing a constraint for subsequent steps. Based on this result, it invokes the license agent to retrieve corresponding licensees (e.g., INTELSAT, GLOBECOMM) via SQL queries, grounding the task in structured data. Finally, the proceeding agent performs graph-based retrieval to compare these entities’ positions on spectrum sharing and reveal consistent opposition among incumbents. At each step, intermediate results are incorporated into the task state, allowing the supervisor to adaptively refine subsequent actions and resolve inter-step dependencies. This example demonstrates how \alg supports coherent cross-source reasoning through dynamic coordination of heterogeneous retrieval mechanisms.

\subsection{Failure Analysis}

To better understand the small set of remaining non-optimal cases, we manually inspect queries where SpecMind does not achieve the best result or produces a partially complete answer. We observe three recurring edge conditions. \textbf{(i) Fine-grained attribution.} In some proceeding graphs, entity--filing relations indicate co-occurrence but do not always explicitly encode whether the entity authored the filing or was referenced by another party, which can make fine-grained attribution more challenging. \textbf{(ii) Sparse evidence for long-tail entities.} Stakeholders that appear only a few times in a proceeding naturally have sparser graph representations, and retrieval for such entities may surface broader contextual information rather than the most specific statement of position. \textbf{(iii) Ambiguity in structured queries.} Some identifiers are only unique within a service-specific subtable, so queries that omit the service type can admit multiple reasonable interpretations and lead to broader aggregation than intended. Overall, these cases largely reflect edge conditions in evidence representation and query specification rather than systematic limitations of the coordination framework. They also point to natural extensions, including richer authorship metadata in graph construction and lightweight clarification mechanisms for ambiguous queries.

\section{Conclusions}

In this paper, we introduced {\alg}, a multi-agent hybrid RAG framework for heterogeneous spectrum data, together with {\algg}, a 450-question benchmark spanning licenses, regulations, proceedings, and cross-source tasks. The results show that {\alg} is particularly effective when conventional semantic retrieval is poorly matched to the underlying data structure: SQL-based retrieval is critical for license records, graph-based retrieval benefits multi-document proceedings, and their coordination yields substantial gains on compound queries. In contrast, the smaller improvement on regulation questions suggests that conventional RAG remains competitive when the source is already well suited to semantic retrieval. These findings indicate that the value of {\alg} lies in selecting and composing retrieval mechanisms according to the structure of the evidence rather than relying on a single universal retrieval pipeline.

Our analysis further characterizes the remaining non-optimal cases and highlights opportunities for refinement, particularly in fine-grained evidence representation and query disambiguation. Future work can build on these directions while expanding the scale and coverage of spectrum policy datasets to support more comprehensive evaluation across broader regulatory domains, frequency bands, and stakeholder interactions. To facilitate such studies, we publicly release {\algg}, the reusable spectrum databases, and the {\alg} framework, providing a common basis for evaluating retrieval-augmented reasoning over heterogeneous spectrum data\cite{RAG_in_Spectrum_Policy}.


\section*{Acknowledgment}

The authors would like to thank Zhiyu Shen, Yuxi Chen, Omkar Mujumdar, Yankai Peng, Hasan Nazim Bicer, Christopher Wahl, Solbee Kang, and Lucas Scholler in the Notre Dame Wireless Institute for their valuable contributions to data collection and the evaluation process.

In addition, we thank Caleb Reinking, Le Li Kruczek, Connor Howington and  Paul Brenner in the Notre Dame Center for Research Computing for their contributions to code optimization and the implementation of the system user interface. Last but not the least, this project was conceived during a SpectrumX center meeting, and subsequently received a seed fund to carry out the research. We thank NSF and SpectrumX for the support.

\newpage
\bibliographystyle{IEEEtran}
\bibliography{references}

\end{document}